\documentclass[10pt,twocolumn,letterpaper]{article}

\usepackage[pagenumbers]{iccv} 

\usepackage{multirow}
\usepackage{makecell}
\usepackage{tikz}
\usepackage{alphalph}
\usepackage{pifont}

\def\methodshort{LOTS\xspace}
\def\method{LOcalized Text and Sketch for fashion image generation\xspace}
\def\dataset{Sketchy\xspace}
\def\suppmat{\textit{Supp. Mat.}\xspace}

\definecolor{block-gray}{gray}{0.96}
\definecolor{zshot-blue}{RGB}{218,232,252}
\definecolor{ft-red}{RGB}{248,206,204}
\definecolor{lots-ca}{RGB}{203, 195, 227}
\definecolor{method-green}{RGB}{213,232,212}

\newcommand{\inlineColorbox}[2]{\begingroup\setlength{\fboxsep}{1pt}\colorbox{#1}{\hspace*{2pt}\vphantom{Ay}#2\hspace*{2pt}}\endgroup}
\usepackage{subcaption}

\usepackage[most]{tcolorbox}
\newtcolorbox{codeblock}{colback=block-gray,grow to right by=-2mm,grow to left by=-2mm,
boxrule=0pt,boxsep=0pt}

\usepackage{amsthm,amsmath,amssymb,amsfonts,bm,bbm,stmaryrd}

\def\eqref#1{equation~\ref{#1}}

\def\1{\bm{1}}

\DeclareMathAlphabet{\mathsfit}{\encodingdefault}{\sfdefault}{m}{sl}
\SetMathAlphabet{\mathsfit}{bold}{\encodingdefault}{\sfdefault}{bx}{n}

\newcommand{\range}[2]{#1..#2}

\definecolor{iccvblue}{rgb}{0.21,0.49,0.74}
\usepackage[pagebackref,breaklinks,colorlinks,allcolors=iccvblue]{hyperref}

\def\paperID{10252} 
\def\confName{ICCV}
\def\confYear{2025}

\title{LOTS of Fashion! \\
Multi-Conditioning for Image Generation via Sketch-Text Pairing}

\author{
Federico Girella\textsuperscript{$1$} 
\quad Davide Talon\textsuperscript{$2$} 
\quad Ziyue Liu\textsuperscript{$1, 3$} 
\quad Zanxi Ruan\textsuperscript{$1$} \\
\quad Yiming Wang\textsuperscript{$2$} 
\vspace{5pt}
\quad Marco Cristani\textsuperscript{$1,4$} \\
\textsuperscript{$1$}University of Verona, 
\textsuperscript{$2$}Fondazione Bruno Kessler, \\
\vspace{10pt}
\textsuperscript{$3$}Polytechnic Institute of Turin,
\textsuperscript{$4$}University of Reykjavik \\
\url{https://intelligolabs.github.io/lots}
}

\begin{document}
\twocolumn[{%
    \renewcommand\twocolumn[1][]{#1}%
    \maketitle
    \thispagestyle{empty}
    \vspace{-8mm}
    \begin{center}
    \captionsetup{type=figure}
     \begin{tikzpicture}
        \node[anchor=south west,inner sep=0] (image) at (0,0) {\includegraphics[width=1.0\linewidth]{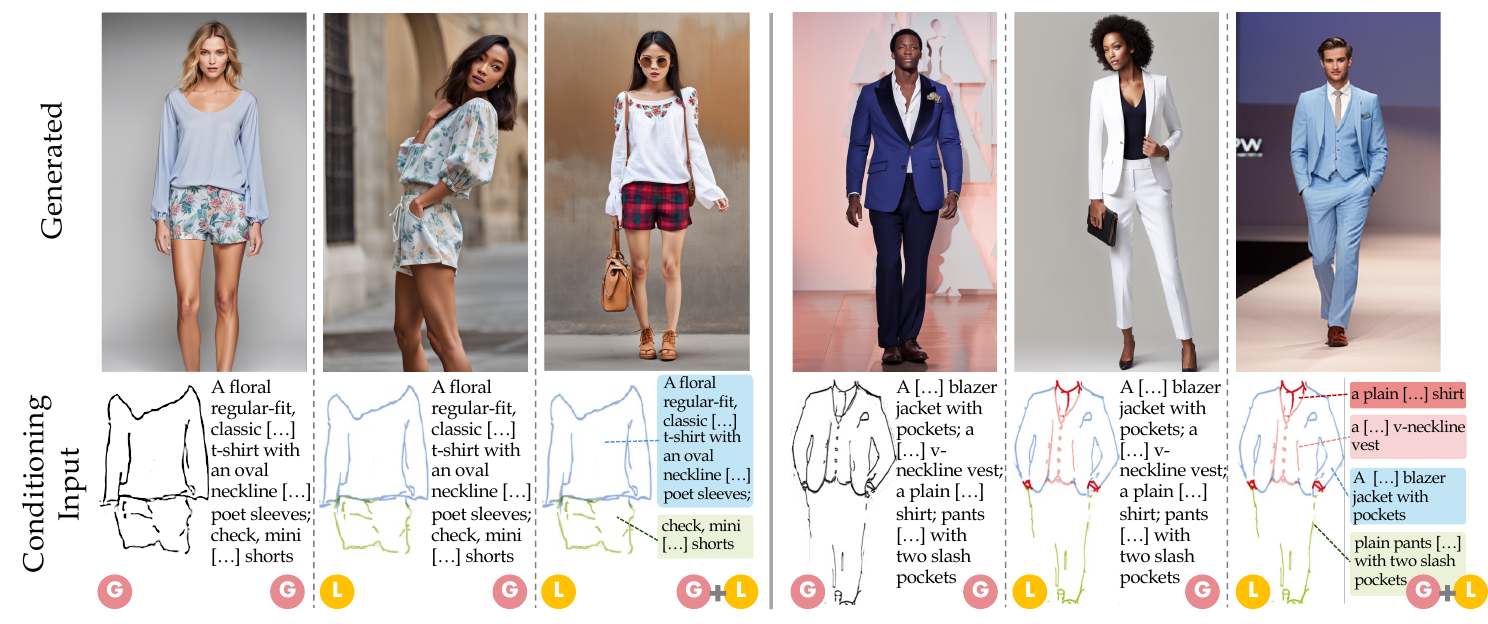}};
        \begin{scope}[x={(image.south east)},y={(image.north west)}]
            \node at (0.135,1.02) {\textbf{IP-Adapter}\cite{ye2023ip}};
            \node at (0.287,1.02) {\textbf{Multi-T2I}\cite{mou2024t2i}};
            \node at (0.432,1.02) {\textbf{Ours}};
            \node at (0.6,1.02) {\textbf{IP-Adapter}\cite{ye2023ip}};
            \node at (0.752,1.02) {\textbf{Multi-T2I}\cite{mou2024t2i}};
            \node at (0.897,1.02) {\textbf{Ours}};
        \end{scope}
    \end{tikzpicture}
     \caption{We present LOTS, enabling fashion image generation with an unprecedented level of control. LOTS represents the natural evolution of fashion design methodologies, progressing from global text and sketches (IP-Adapter~\cite{ye2023ip}) to localized sketches with global text (Multi-T2I~\cite{mou2024t2i}). Our approach leverages a global description (omitted here for brevity) alongside a set of localized sketch-text pairs (the coloured boxes), effectively defining both the layout and appearance of individual garment items.}
\label{fig:teaser}
\end{center}}]

\begin{abstract}
Fashion design is a complex creative process that blends visual and textual expressions. Designers convey ideas through sketches, which define spatial structure and design elements, and textual descriptions, capturing material, texture, and stylistic details.
In this paper, we present \method{} (\methodshort{}), an approach for compositional sketch-text based generation of complete fashion outlooks. \methodshort{} leverages a global description with paired localized sketch + text information for conditioning and introduces a novel step-based merging strategy for diffusion adaptation. 
First, a Modularized Pair-Centric representation encodes sketches and text into a shared latent space while preserving independent localized features; then, a Diffusion Pair Guidance phase integrates both local and global conditioning via attention-based guidance within the diffusion model’s multi-step denoising process. 
To validate our method, we build on Fashionpedia to release \dataset{}, the first fashion dataset where multiple text-sketch pairs are provided per image. Quantitative results show \methodshort{} achieves state-of-the-art image generation performance on both global and localized metrics, while qualitative examples and a human evaluation study highlight its unprecedented level of design customization.
\end{abstract}    
\section{Introduction}
\label{sec:intro}

In fashion design, designers need to express their abstract inspirations through forms that are natural to humans, \eg, sketches or natural language. 
For example, a designer can sketch the shape of a t-shirt, with its design idea expressed in natural language description ``\textit{a floral, regular-fit, classic t-shirt with an oval neckline and wrist-length poet sleeves}'' or the shorts as \textit{``check, mini, symmetrical, gathering shorts with a regular fit''}, as shown in Fig.~\ref{fig:teaser} (left). 
Sketches and natural language descriptions associated with the same garment convey complementary information for depicting the final design: sketches provide spatial information for the garment outline, such as its fit (\eg, the tight fit jacket in Fig.~\ref{fig:teaser} (right)), while the language description conveys richer details in terms of accessories and styles (\eg, ``a blazer jacket with a pocket'').
Since a complete design is composed of several clothing garments, multiple natural descriptions are often collected together to outline an outfit. Each description specifies a \textit{localized part} of the design, in terms of silhouette shapes, materials, and textual details, allowing fine-grained localized control over the generation. 

In this paper, we aim to facilitate the process of concretizing creative ideas into graphical output, as expressed in the form of localized conditions, \ie, sketch-text pairs, to support the designing process for the fashion industry.
We frame this problem as a conditional image generation task, where the conditioning consists of a set of localized text-sketch pairs.
In addressing such multi-localized-conditioning generation, the state-of-the-art methods fall short. 
Recent diffusion adapters for sketch-to-image generation~\cite{mou2024t2i, ye2023ip, sun2024anycontrol, zhang2023adding} allow for multi-spatial conditions but underperform when providing fine-grained textual information, such as neckline type, pocket/button style, as shown in \cref{fig:teaser} (Multi-T2I~\cite{mou2024t2i}). 
We argue that this limitation stems from the use of a \textit{single global description} to inject textual condition: all relevant details about different parts of the garment are considered in a monolithic fashion, leading to incorrect localization of attributes to parts~\citep{mou2024t2i, ye2023ip, sun2024anycontrol, zhang2023adding}. 
We refer to this problem as ``attribute confusion'', where properties of one item are instead generated for another item, \eg, ``a floral t-shirt with check short'' could bring the floral pattern in the shorts Fig.~\ref{fig:teaser} (left). 

We propose \method{} (\methodshort{}), a novel approach leveraging multiple localized sketch-text pairs for image conditioning. First, the \textit{Modularized Pair-Centric Representation} module independently encodes sketches and text into a shared latent space, preserving localized features while limiting information leakage between pairs to reduce attribute confusion. Then, the \textit{Pair-former} merges text and sketch information within each pair, ensuring spatially grounded descriptions that capture single-item attributes through the structural guidance of sketches.
Next, during the \textit{Diffusion Pair Guidance} phase, these localized representations serve as conditioning inputs to a pre-trained diffusion model, alongside a global textual representation specifying general appearance properties (style, background). 
Unlike prior methods,
our approach defers 
condition inputs merging
to the diffusion process itself, breaking down the task across multiple denoising steps via a cross-attention strategy. 

For model training and comparative evaluation, we introduce \dataset{}, a new dataset built on Fashionpedia~\citep{jia2020fashionpedia} for localized sketch-to-image generation. In this dataset, garments within an outfit are enriched with sketches and treated as multiple conditioning inputs, each paired with fine-grained descriptions. \methodshort{} achieves state-of-the-art performance in image quality
and attribute localization, as evidenced by quantitative metrics (+3.4\% GlobalCLIP \textit{wrt} fine-tuned ControlNet), human evaluation studies (+3.1\% F1 Score over SDXL), and qualitative analysis. 

\noindent{\textbf{Our contributions}} are four-fold:
\begin{itemize}
\item We focus on localized sketch-text image generation, advancing state-of-the-art conditioning with localized sketches and a single global text. 

\item We introduce a novel method, \methodshort{}, to mitigate attribute confusion by modularized attention-based processing per sketch-text pair, and deferring the multi-conditioning merge to the denoising process.

\item We introduce a new dataset, \dataset{}, in the fashion domain, to facilitate model training and evaluation for the localized text-sketch image generation problem. 

\item \methodshort{} achieves state-of-the-art performance in sketch-text conditioning and attribute localization, as measured with both metrics and human evaluation.

\end{itemize}

\section{Related Works}
\label{sec:related}
In this section, we focus on text-to-image, sketch-to-image, and controllable diffusion-based generation. 

\noindent\textbf{Text-to-Image Generation.}
Recent advances in Text-to-Image (T2I) generation have been driven by diffusion models~\cite{ho2020denoising,ho2022classifier,song2020denoising}, which generate high-quality images from textual prompts~\cite{nichol2021glide,rombach2022high,ramesh2022hierarchical,saharia2022photorealistic} through a noise-based forward and denoising reverse process. Conditioning techniques enhance control: GLIDE~\cite{nichol2021glide} employs classifier-free guidance, DALLE-2~\cite{ramesh2022hierarchical} uses a two-stage CLIP-based approach, and Imagen~\cite{saharia2022photorealistic} integrates large-scale language models for improved realism and semantic alignment. Stable Diffusion (SD)~\cite{rombach2022high} refines conditioning via cross-attention while optimizing efficiency through latent-space diffusion.
Building on SD-like models, we extend control beyond textual descriptions, focusing on multiple sketch-text pairs for localized and fine-grained conditioning.

\noindent\textbf{Sketch-to-Image Generation.}
Sketch-to-Image generation has evolved from GAN-based methods~\citep{isola2017image, lu2018image, ghosh2019interactive, koley2023picture, richardson2021encoding} to pre-trained diffusion models~\citep{wang2022pretraining, voynov2023sketch, mengsdedit}. On this line,  
PITI~\cite{wang2022pretraining} maps sketches to semantic latents of a large-scale diffusion model, while SDEdit~\citep{mengsdedit} employs sketch perturbation and denoising to guide generation. Finally, LGP~\citep{voynov2023sketch} enforces consistency between noisy features and spatial sketch guidance.
Recent approaches for sketch-to-image generation have implemented different techniques to control the downstream diffusion model that address the problem of spatial conditioning~\citep{zhang2023adding, mou2024t2i, ye2023ip}, sketch abstraction~\citep{navard2024knobgen, koley2024s}, or professional sketches~\citep{wang2024lineart}. Unlike prior work~\citep{zhang2023adding, mou2024t2i, ye2023ip, navard2024knobgen, koley2024s, wang2024lineart, wang2022pretraining, mengsdedit, voynov2023sketch}, which conditions on global sketches, we introduce localized sketch control.

\noindent\textbf{Controllable diffusion-based generation.}
While textual prompts enable high-quality image generation in T2I models, they often lack fine-grained control. To enhance controllability, various methods integrate additional conditioning elements~\cite{zhang2023adding, huang2023t2i, ye2023ip, li2023gligen, sun2024anycontrol, zhao2024uni}, including bounding boxes~\citep{li2023gligen}, blobs~\citep{nie2024compositional}, and segmentation masks~\citep{kim2023dense, goel2024pair}.
GLIGEN~\citep{li2023gligen} conditions the diffusion model with bounding box coordinates to localize textual information, but does not allow for paired sketch-text localization.
ControlNet~\citep{zhang2023adding} introduces zero-convolution modulation in a frozen diffusion model, while subsequent works propose multi-modal control~\cite{hu2023cocktail} and all-in-one control adapters~\cite{zhao2024uni}, though both rely on fixed-length input channels. AnyControl~\cite{sun2024anycontrol} enables multi-control conditioning but requires a trainable copy of the diffusion model. Alternative methods, such as T2I~\citep{mou2024t2i} and IP~\citep{ye2023ip} adapters, employ residual feature maps and cross-attention, respectively, to aggregate multiple control signals before conditioning. However, these approaches depend on global textual prompts and are limited by the 77-token constraint of text encoders. In contrast, we couple localized textual descriptions with their corresponding sketches to improve fine-grained generation. Our adapter enables a pre-trained T2I diffusion model to condition on a variable number of sketch-text pairs while remaining lightweight to train.
Furthermore, recent approaches~\citep{Baldrati_2023_ICCV,xie2025hierafashdiff} employ image editing in the Fashion domain. Multimodal Garment Designer\citep{Baldrati_2023_ICCV} requires a starting image as input for the edit, while \methodshort{} performs image generation from scratch. HieraFashDiff~\citep{xie2025hierafashdiff} is a concurrent work presenting a two-stage pipeline performing generation and iterative editing. Differently, \methodshort{} performs one-shot generation and allows for additional sketch conditioning. Since direct comparison would require non-trivial modifications, risking unfair or misleading results, we will not compare to editing-based approaches.

\section{Method}
\label{sec:method}

\begin{figure*}
    \centering
    \includegraphics[width=1.0\linewidth]{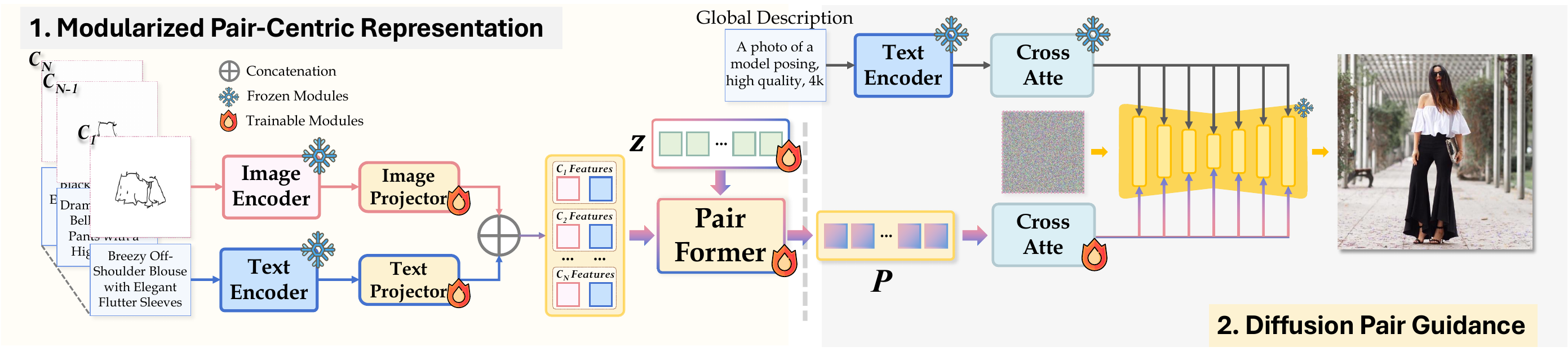}
    \caption{\methodshort{} mitigates attributes confusion building on paired sketch-text conditioning for image generation. \textbf{1.} In an initial phase, the modularized Pair-Centric Representation (Sec.~\ref{sec:pair-centric-representation}) independently processes available pairs by first embedding the different modalities with pre-trained modality-specific encoders, and later localizing the semantic textual information according to the associated sketch structure in the Pair-Former. \textbf{2.} In the second Diffusion Pair Guidance phase (Sec.~\ref{sec:diffusion-guidance}), pair representations are directly injected into the downstream diffusion model. By breaking down the merge task within the denoising diffusion steps, \methodshort{} avoids explicit merge of pair representations that lead to attribute confusion.}
    \label{fig:method}
\end{figure*}

In this section we present \methodshort{}: \emph{\method{}}.
We start with a formalization of the task, defining the input-output of the problem, and later introduce the key modules of our adapter strategy.

\noindent\textbf{Localized sketch-to-image generation.} 
The input and desired output of the problem are as follows.
\textbf{Input:} Let $\mathcal{C}$ be a set of sketch-text pairs $\mathcal{C}=\{C_1, .., C_{N_i}\}$ where $C_i = (S_i, T_i)$ denotes the $i$-th sketch-text pair with sketch conditioning $S_i \in \{0, 1\}^{H\times W}$ and textual description $T_i$, with $N_i$  the number of pairs associated with the $i$-th sample and $H, W$ denoting the width and height of the desired generated image, respectively. We assume that the provided sketches have a globally coherent spatial layout. In other words, local sketches associated with an item should satisfy the constraint $\sum_i^N S_i = S$, where $S$ is the global sketch of the desired image. To enable global information conditioning in natural language, we further allow for the global textual representation $T_g$ to prompt the model with general appearance information that the desired image should have, \eg, the fashion style or the background. \textbf{Output:}  Localized sketch-to-image generation aims to synthesize with the generative model $\phi$ an image $X\in\mathbb{R}^{3\times H \times W}$ based on the conditioning input $\mathcal{C}$ as $X = \phi(\mathcal{C}, T_g)$.
The resulting generation should faithfully preserve the spatial constraints of sketches and the semantic details of text descriptions while ensuring global coherence 
between all localized pairs and global description.
In particular, the specified properties of the textual description $T_i$ associated with the $i$-th item should be correctly reflected in the
location described by its associated spatial information $S_i$,
while not leaking to other local parts of the image $S_j$, with $i,j =\range{1}{N}, j\neq i$.

\noindent\textbf{Method overview.}  \methodshort enables sketch-text conditioned image generation through a two-phase process. 
An illustration of the proposed approach is presented in Fig.~\ref{fig:method}. 
In the \textit{Modularized Pair-Centric Representation} phase (Section~\ref{sec:pair-centric-representation}), sketch and text inputs are independently encoded using pre-trained modality-specific encoders and then projected into a multi-modal shared latent space via the \textit{Pair-Former module}, ensuring localized feature extraction without interference between pairs. To address \textit{attribute confusion} when merging multiple conditioning inputs, the \textit{Diffusion Pair Guidance} phase (Section~\ref{sec:diffusion-guidance}) employs attention-based conditioning within the multi-step denoising process, 
integrating both local and global conditioning effectively.

\subsection{Modularized Pair-Centric Representation}
\label{sec:pair-centric-representation}
Localized sketch-to-image generation requires generating an image starting from the set of local conditionings $\mathcal{C} = \{C_1,.., C_2\}$ while ensuring no semantic information from $C_i$ is leaking to unrelated parts of the image.
To address this, we propose to independently process the input information in a pair-centric fashion: each pair is considered 
independent,
or in other words, pairs do not influence or see each other. Consider a single localized representation in the form of a sketch-text pair $C_i = (S_i, D_i), i=\range{1}{N}$. We embed conditioning information via modality-specific encoders:
\begin{align}
    h^T_i &= f^T(T_i)\\
    h^S_i &= f^S(S_i)
\end{align}
where $f^T$ and $f^S$ denote respectively the text and sketch encoders, and $h^T_i, h^S_i$ their associated output latent representations, $i=\range{1}{N}$. 
\paragraph{Pair-Former.}\label{subsec:pairformer} Starting from the modality-specific representations, for each pair, we integrate the sketch spatial guidance and the text semantic information into a shared feature space, where textual information is localized according to the sketch structure.
Inspired by recent vision-language advancements~\citep{li2023blip}, we devise a new self-attention strategy aiming to (i) compress the sparse sketch representation in a limited number of tokens and (ii) merge multi-modal information. Let $z \in \mathbb{R}^{k \times d}$ be a set of $k$ $d$-dimensional learnable tokens prepended to the concatenation of visual and textual embedding representations. We obtain the pair tokens by first computing the self-attention with sketch and text representations:
\begin{equation}
 h_i = \texttt{Self-attn}([z; h^S_i; h^T_i]),
\end{equation}
and restricting $p_i$ to be the first $k$ tokens of $h_i$, \ie, the output tokens associated to $z$,
where $[\cdot\ ;\ \cdot]$ denotes the concatenation operation along the token dimension. Hence, self-attention allows the prepended token $z$ to pool the informative content from both the sketch and textual modalities in a coordinated manner. We remark that while the $z$ learnable tokens are shared between the different pairs, the Pair-former is independently processing the $N$ input pairs.

\subsection{Diffusion Pair Guidance}
\label{sec:diffusion-guidance}
Merging information from multiple conditioning pairs is challenging, as it requires effective coordination while avoiding information leakage. Existing approaches encode multiple guidance signals in a single pooling step, which may cause interference among pairs.
In contrast, we defer the merging process to the pre-trained diffusion model, leveraging its iterative denoising steps to progressively integrate conditioning pairs, rather than combining them in a single step. Given $P\in \mathbb{R}^{N\times k \times d}$ as the concatenated sequence of the $N$ pooling tokens $p_i, i=\range{1}{N}$ extracted by our Pair-Former, we avoid explicitly merging the pairs. Instead, we inject the entire sequence into the diffusion model, allowing the merge operation to occur throughout the entire diffusion process. This allows for more dynamic interaction between the multiple given conditionings during the reverse diffusion process.
Inspired by previous work \cite{ye2023ip}, we rely on cross-attention layers to guide diffusion generation and introduce an additional learnable cross-attention layer $\hat{w}$ after each pre-existing cross-attention layer $w$ in the frozen denoising model. These new layers inject the conditioning sequence $P$ into the model features at each diffusion timestep, allowing for the iterative merging of information. Let $x$ denote the input features of a global text-conditioning cross-attention layer in the denoising network, the conditioned output $x'$ of the paired cross-attention layers is computed as: 
\begin{equation}
x' = w(x, h^{T_g}) + \alpha \hat{w}(x, P), \end{equation} 
where $w(\cdot, \cdot)$ represents the cross-attention between the two input token sequences and $h^{T_g}$ denotes the embedding representation of the global text prompt guiding the model with semantic information that should be globally represented, \eg, the style or the background. Here, $\alpha$ is a scaling hyper-parameter, constrained to the range $[0, 1]$, that regulates the strength of the guidance from $P$. During training, we set $\alpha$ to 1 to fully enable the cross-attention layers $\hat{w}$ to learn the appropriate merging behavior.
Notably, the attention-based nature of these adapter blocks allows an arbitrary number of conditioning tokens, enabling \methodshort to work with a variable number of conditioning pairs.
\section{Experiments}
\label{sec:experiments}
We assess \methodshort{} by comparing its performance with state-of-the-art sketch-to-image adapters in the fashion domain. First, in \cref{subsec:metrics}, we present our newly proposed dataset, detailing its construction and structure, along with the evaluation metrics and implementation details. In \cref{subsec:comparisons}, we outline the experimental setup and baseline models, and analyze the quantitative and qualitative results. Finally, in \cref{subsec:ablations}, we conduct ablation studies to highlight key design choices of \methodshort.

\subsection{The \dataset dataset}
\begin{figure}
    \centering
    \includegraphics[trim={0 11.7cm 16cm 0cm},clip,width=1.0\linewidth]{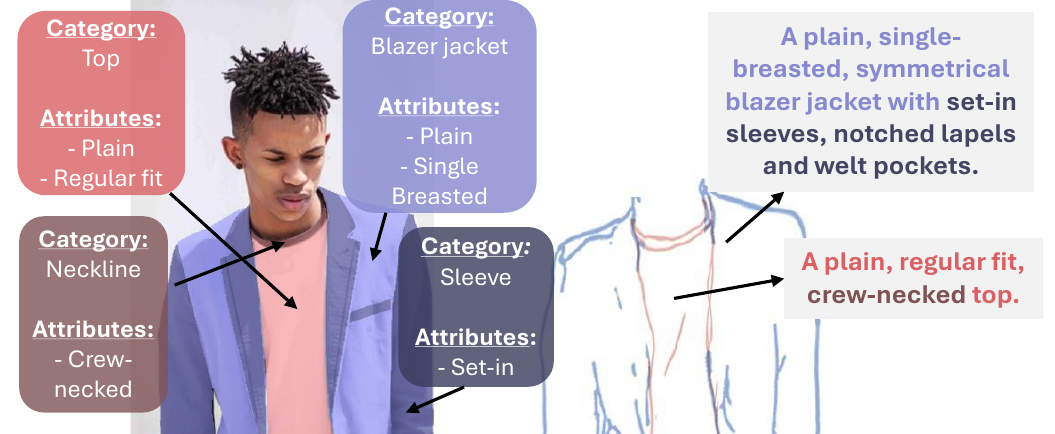}
    \caption{Example of the hierarchical structure of \dataset. Starting from whole-body item (light colors) and garment parts (dark shades) annotations, we build a hierarchical structure by pairing the garment part annotations to their related whole-body garment. Then, we use this structure to generate garment-level sketches and natural language descriptions by relying on off-the-shelf models.}
    \label{fig:dataset}
\end{figure}

We construct a novel dataset, named \dataset, to support training and evaluation on the task of localized sketch-to-image generation, with localized text-sketch pairs associated with high-quality ground truth images. In the following, we explain in detail how we organize the dataset based on garments and the localized text-sketch creation, as well as the dataset statistics.

\noindent\textbf{Local garments organization.} We build \dataset{} on Fashionpedia~\cite{jia2020fashionpedia}, a dataset composed of 46k images for training and 1.2k for testing, where fashion experts annotate garments with fine-grained attributes and segmentation masks. While these masks include detailed part annotations (\eg, pockets, zippers, sleeves), they lack a hierarchical structure linking garment components. To improve compositionality, we introduce a two-level hierarchical organization based on segmentation mask overlaps.
Specifically, following Fashionpedia taxonomy, the 330k item annotations are first categorized into 14 ``whole-body items'' (\eg, tops, shirts, and skirts) and 32 ``garment parts'' (\eg, sleeves, pockets, and necklines).
To ensure high-quality compositional annotations, we retain all whole-body categories, along with 21 sub-item categories from Fashionpedia, while removing 11 categories (31k annotations) that are rare, seldom overlap, or lack consistent overlap with any whole-body items, such as umbrellas, bags, and glasses.
Then, for each image, we determine the overlap between each garment part's mask and every whole-body item mask. 
Whole-body annotations are considered as top-level annotations, \ie, a garment in the image, while part annotations are assigned to the whole-body item with which they have the greatest overlap, \ie, they are considered sub-garment annotations referring to a property, such as sleeves, necklines, and pockets. 

\noindent\textbf{Localized text-sketch creation.} While the Fashionpedia annotations are rich in attributes, they lack a coherent natural language description, which is essential for our text conditioning. Thus, we generate the textual description for each garment in the image by prompting a pre-trained Large Language Model~\cite{touvron2023llama} with the hierarchical annotation structure of each garment, along with some in-context learning examples of the desired reply format.
We further augment the dataset by including localized garment-level sketches, generated from the ground-truth images, using a pre-trained Image-to-Sketch model~\cite{li2019photo}. We remove background information via masking to ensure each sketch contains only information about the associated item.
We provide a global composition of all the garment sketches, which depicts the sketch of the entire outfit in the original image.
Finally, we pre-process the images by resizing them to 512 pixels. We preserve their aspect ratio with white padding into a square format to maintain consistency across samples. Additional details are available in the \suppmat.

\noindent\textbf{Dataset statistics.} Our \dataset extends Fashionpedia, providing a total of 47k images and 79k garment-level annotations, resulting in an average of 1.7 annotations per image (min 1, max 6). 
As shown in~\cref{fig:dataset}, each annotation contains the associated sketch, hierarchical attributes, and natural language description of the item. The average word length of the descriptions is 16 words. 

\subsection{Experimental protocol}
\label{subsec:metrics}
\noindent\textbf{Quantitative evaluation and metrics.} 
Following prior works~\cite{sun2024anycontrol,goel2024pair,lukovnikov2024layout,cheng2023adaptively,bashkirova2023masksketch}, we adopt Fréchet Inception Distance (FID)~\cite{heusel2017gans}
to assess the global fidelity of the generated images distribution, relative to ground truth images. Lower values of FID indicate better perceptual quality and stronger correspondence.
To measure the semantic alignment, we rely on GlobalCLIP~\citep{radford2021learning} score, calculated as the cosine similarity of image embeddings,  in line with~\cite{bashkirova2023masksketch}.
To further capture fine-grained and localized alignment, we adopt LocalCLIP score, building on~\cite{lukovnikov2024layout, kim2023dense}. Specifically, we first use the masking annotations to obtain a crop for each garment in the ground truth and generated images. Then, we apply the CLIP~\citep{radford2021learning} visual encoder to compute the cosine similarity between each pair of ground truth and generated image crops, averaging their score across all garments. Higher scores correspond to better semantic alignment.
In addition, we utilize the state-of-the-art compositional semantic alignment metric, VQAScore~\cite{lin2024evaluating} that captures how well-generated images reflect complex text descriptions by relying on Visual Question Answering approaches to query the presence of desired properties. A larger VQAScore suggests improved compositional alignment to the provided prompt.
Finally, we follow previous work~\cite{goel2024pair,wang2024lineart} and evaluate the proposed \methodshort{} based on the Structural Similarity Index Measure (SSIM) assessing the sketch structural alignment and edge fidelity (the higher, the better).

\noindent\textbf{Human evaluation.}
In line with prior works~\cite{zhang2023adding,zhao2024uni}, 
we conduct a user study involving 14 participants with an average of 57 answers each, targeting the evaluation of localized control, rather than visual perception.
We generate images ensuring that attributes appear only once in the entire outfit. We leverage a questionnaire to assess whether a specific attribute associated with the $i$-th garment is correctly localized in the desired garment of the image and if it leaks to other ones. 
We quantitatively evaluate considered models in terms of Precision ($\uparrow$), Recall ($\uparrow$), and F1 Score ($\uparrow$) metrics with respect to localized conditioning. 
A high F1 Score indicates the model's overall performance in correctly reflecting and localizing the attribute, showing its capacity to balance accurate placement with less attribute confusion. Additional details are available in the \suppmat

\subsection{Main Comparisons}\label{subsec:comparisons}
\noindent\textbf{Baselines.} We compare \methodshort with baselines and state-of-the-art sketch-to-image approaches.
Specifically, for text-only approaches, we evaluate Stable Diffusion 1.5 (SD)~\cite{rombach2022high} and Stable Diffusion XL (SDXL)~\cite{podell2023sdxl}, which generate images solely from a global text prompt,
and GLIGEN~\citep{li2023gligen}, which enables localized textual conditioning.
We then compare to sketch-to-image adapter methods, including SD-based ControlNet~\cite{zhang2023adding}, SDXL-based T2I-Adapter~\cite{mou2024t2i} and SDXL-based IP-Adapter~\citep{ye2023ip}, which all incorporate both a global text prompt and a global sketch as inputs, offering enhanced spatial conditioning.
Additionally, we examine compositional modifications of ControlNet (Multi-ControlNet) and T2I-Adapter (Multi-T2I-Adapter) that were modified to allow multiple local sketches conditioning alongside a single global text prompt.
Furthermore, we evaluate the most recent localized control method, AnyControl~\citep{sun2024anycontrol}, that allows for local sketches and a single global text description as inputs.
Finally, we also assess the performance of the adapter-based approaches with fine-tuning on our \dataset dataset.

\noindent\textbf{Experimental setup.}
We adapt conditioning to different generative models based on their input requirements. For global descriptions, we concatenate all garment descriptions, while for models requiring a single image guide, we create a composite global sketch by merging individual garment sketches. When localized control is supported, we use garment-specific sketches and/or descriptions as input.
For consistency, all images are generated at \texttt{(512x512)} resolution using each model’s default inference setup. Additionally, to ensure fairness, our global description $T_g$ remains fixed across samples as ``A picture of a model posing, high-quality, 4k''. Experiments exploring variations in global descriptions are detailed in 
the \suppmat

\noindent\textbf{Implementation Details.} We adopt \texttt{DINOv2 vits14}~\cite{oquab2023dinov2} as sketch encoder. As text encoder, we follow the findings in~\cite{podell2023sdxl} and use a combination of \texttt{OpenCLIP ViT-bigG}~\cite{openclip} and \texttt{CLIP ViT-L}~\cite{radford2021learning} by concatenating the penultimate text encoder outputs along the channel-axis. Additional implementation details are available in the \suppmat

\noindent\textbf{Quantitative results.} 
Tab.~\ref{tab:vlm_stats} reports the quantitative evaluation across global quality, semantic alignment, and structural similarity metrics on the test split of \dataset. Our method demonstrates state-of-the-art performance in GlobalCLIP, LocalCLIP, and VQAScore, while ranking third in SSIM, indicating overall superiority and strong alignment both semantically and structurally.
Specifically, \methodshort attains the second lowest FID, demonstrating a high perceptual fidelity. 
A similar pattern is observed in GlobalCLIP and LocalCLIP scores, with our method surpassing all baselines and state-of-the-art models (+3.4\% and +1.2\%), indicating strong semantic alignment and feature similarity with the ground truth.
For compositional semantic alignment, T2I-Adapter and our method achieve the highest VQAScore, outperforming all alternatives in textual prompt following.
While some models outperform \methodshort in some metrics, their improvements come with trade-offs: the fine-tuned IP-Adapter achieves a lower FID but sacrifices both text and sketch guidance, as evidenced by its significantly lower Compositional Alignment metrics.
Similarly, for structural similarity, Multi-T2I-Adapter achieves the highest SSIM, followed by IP-Adapter, while our method ranks third. As opposed to \methodshort, however, both IP-Adapter and Multi-T2I-Adapter emphasize sketch-guidance over prompt adherence and image coherence, as evidenced by their subpar LocalCLIP, VQAScore, and FID scores, which figure among the lowest.
In conclusion, these results highlight how, thanks to its novel pairing strategy, \methodshort strikes an optimal balance between image quality and prompt adherence, 
surpassing prior approaches and 
setting the new state-of-the-art performance in the fashion localized sketch-to-image task.

\begin{table*}
    \begin{minipage}[t]{0.62\linewidth}
    \centering
    \resizebox{\linewidth}{!}{
    \begin{tabular}{l c ccc ccc}
        \toprule
        \multirow{2}{*}{\textbf{Model}} & \multirow{2}{*}{\textbf{Conditioning}}& \multicolumn{2}{c}{\textbf{Global Quality}} 
        &\multicolumn{3}{c}{\textbf{Compositional Alignment}}\\ 
        \cmidrule(lr){3-4}
        \cmidrule(lr){5-7}
         & Visual/Textual & FID ($\downarrow$) & GlobalCLIP ($\uparrow$) & LocalCLIP ($\uparrow$) & VQAScore ($\uparrow$) & SSIM ($\uparrow$)\\
        \midrule
        \cellcolor{zshot-blue}{SD~\cite{rombach2022high}} & -/G & 1.11 & .603 & .745 & .719 & .663\\
        \cellcolor{zshot-blue}{SDXL~\cite{podell2023sdxl}} & -/G & 1.77 & .529 & .701 & .660 & .544\\
        \cellcolor{zshot-blue}{GLIGEN~\cite{li2023gligen}} & -/L & 0.93 & .568 & .704 & .395 & .614\\
        \cellcolor{zshot-blue}{ControlNet~\cite{zhang2023adding}} & G/G & 1.08 & .622 & .789 & .733 & .674\\
        \cellcolor{zshot-blue}{Multi-ControlNet~\cite{zhang2023adding}} & L/G & 1.10 & .615 & .780 & .730 & .672\\
        \cellcolor{zshot-blue}{IP-Adapter~\cite{ye2023ip}} & G/G & 2.80 & .537 & .682 & .611 & \underline{.715}\\
        \cellcolor{zshot-blue}{T2I-Adapter~\cite{mou2024t2i}} & G/G& 2.16 & .534 & .705 & .635 & .482\\
        \cellcolor{zshot-blue}{Multi-T2I-Adapter~\cite{mou2024t2i}} & L/G& 1.14 & .583 & .766 & .697 & \textbf{.723}\\
        \cellcolor{zshot-blue}{AnyControl~\citep{sun2024anycontrol}} & L/G & 0.99 & .602 & .777 & .712 & .544\\
        \midrule
        \cellcolor{ft-red}{GLIGEN~\cite{li2023gligen}} & -/L & 1.70 & .564 & .713 & .419 & .514\\
        \cellcolor{ft-red}{ControlNet~\cite{zhang2023adding}} & G/G & 0.80 & \underline{.645} & \underline{.801} & .717 & .574\\
        \cellcolor{ft-red}{Multi-ControlNet~\cite{zhang2023adding}} & L/G & 0.84 & .638 & .799 & .720 & .572\\
        \cellcolor{ft-red}{IP-Adapter~\cite{ye2023ip}} & G/G & \textbf{0.69} & .621 & .787 & .714 & .631\\
        \cellcolor{ft-red}{T2I-Adapter~\cite{mou2024t2i}} & G/G & 1.03 & .570 & .753 & \textbf{.749} & .612\\
        \cellcolor{ft-red}{Multi-T2I-Adapter~\cite{mou2024t2i}} & L/G & 1.11 & .559 & .744 & \underline{.734} & .605\\
        \cellcolor{method-green}\methodshort  \textbf{(Ours)}& L/L & \underline{0.79} & \textbf{.679} & \textbf{.813} & \textbf{.749} & .678\\
        \bottomrule
    \end{tabular}
}

    \subcaption{Comparisons between \methodshort and state-of-the-art sketch-to-image approaches. In the Conditioning column, L and G indicate whether the model accepts Local or Global inputs as Visual or Textual conditioning. We divide the table into three sections: \inlineColorbox{zshot-blue}{zero-shot approaches}, \inlineColorbox{ft-red}{fine-tuned} approaches on \dataset{}, and our approach\inlineColorbox{method-green}{\methodshort} . We highlight the best performance in bold and underline the second best}\label{tab:vlm_stats}
\end{minipage}
\hspace{0.01\linewidth}
\begin{minipage}[t]{0.32\linewidth}
\centering
\resizebox{\linewidth}{!}{%
    \begin{tabular}{l c c c c}
        \toprule
        & \multicolumn{3}{c}{\textbf{Attribute Localization}}\\
        \cmidrule(lr){2-4}
        \textbf{Model} & Precision ($\uparrow$)& Recall ($\uparrow$) & F1 ($\uparrow$)\\ 
        \midrule
        \cellcolor{zshot-blue}{SDXL~\cite{podell2023sdxl}} & .636 & \textbf{.754} & \underline{.690}\\
        \cellcolor{zshot-blue}{ControlNet~\cite{zhang2023adding}} & .596 & .449 & .512\\
        \cellcolor{zshot-blue}{Multi-ControlNet~\cite{zhang2023adding}} & .487 & .365 & .418\\
        \cellcolor{zshot-blue}{IP-Adapter~\cite{ye2023ip}} & .625 & .139 & .227\\
        \cellcolor{zshot-blue}{T2I-Adapter~\cite{mou2024t2i}} & .409 & .170 & .240\\
        \cellcolor{zshot-blue}{Multi-T2I-Adapter~\cite{mou2024t2i}} & .370 & .270 & .312\\
        
        \cellcolor{zshot-blue}{AnyControl~\citep{sun2024anycontrol}} & .281 & .134 & .182\\
        
        \midrule
        
        \cellcolor{ft-red}{ControlNet~\cite{zhang2023adding}} & \underline{.667} & .516 & .582\\
        \cellcolor{ft-red}{Multi-ControlNet~\cite{zhang2023adding}} & .541 & .417 & .471\\
        \cellcolor{ft-red}{IP-Adapter~\cite{ye2023ip}} & .559 & .384 & .455\\
        \cellcolor{ft-red}{T2I-Adapter~\cite{mou2024t2i}} & .463 & .397 & .427\\
        \cellcolor{ft-red}{Multi-T2I-Adapter~\cite{mou2024t2i}} & .551 & .692 & .614\\
        \cellcolor{method-green}\methodshort  \textbf{(Ours)} & \textbf{.813} & \underline{.650} & \textbf{.722}\\
        \bottomrule
    \end{tabular}
    }
    \subcaption{Results of qualitative user study of attribute localization and confusion conducted between \methodshort and other models. We highlight the best results for each metric in bold and underline the second best.}\label{tab:subjective}
\end{minipage}
\caption{On the left, the quantitative metrics on the test split of \dataset. On the right, are metrics obtained through our user study.}
\end{table*}

\begin{figure*}
    \centering
    \includegraphics[width=1.0\linewidth]{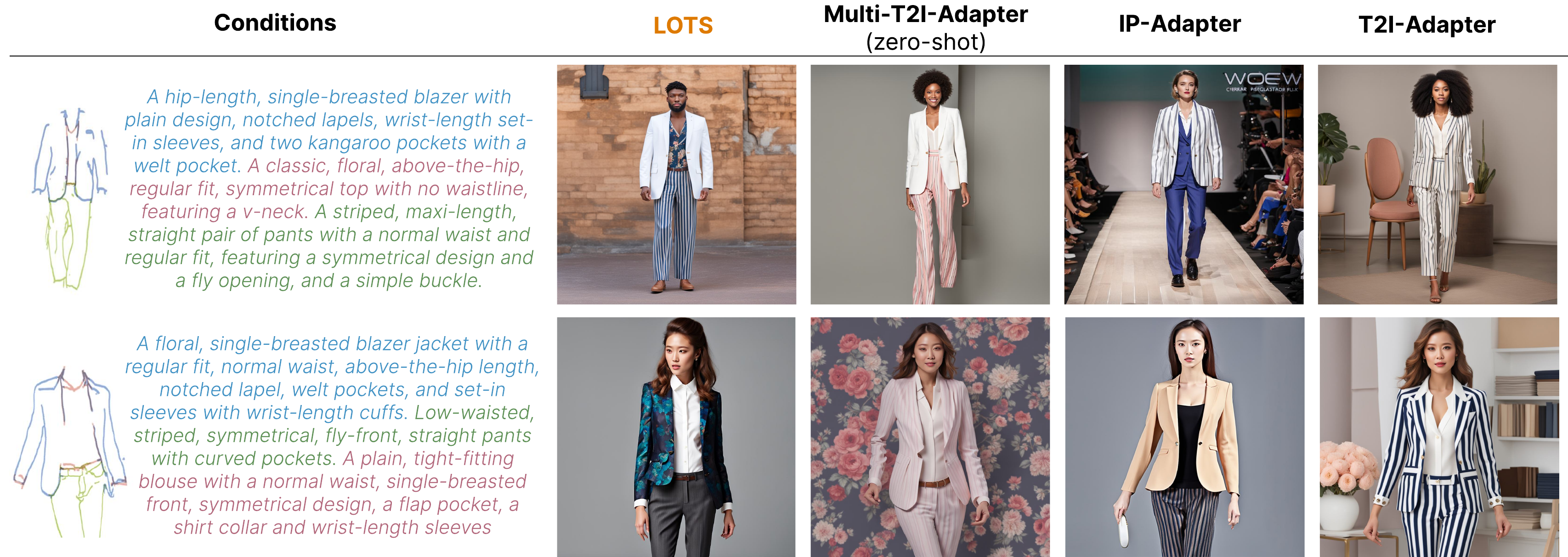}
    \caption{Qualitative results of \methodshort{} in comparison with Multi-T2I-Adapter~\cite{mou2024t2i}, IP-Adapter~\cite{ye2023ip}, and T2I-adapter~\cite{mou2024t2i}. Given paired localized
text-sketch pairs as conditioning inputs, \methodshort{} can better reflect fine-detailed attributes in the intended local region of the generated
images, effectively mitigating attribute confusion.}
    \label{fig:qualitatives}
\end{figure*}

\noindent\textbf{Results with human evaluation.}
In this evaluation, we aim to measure whether an attribute is correctly localized in a garment of the generated image.
In~\cref{tab:subjective}, we present the results of our human study for attribute localization and confusion across different models.
Our \methodshort{} achieves the highest F1 score among all models, indicating superior performance in accurately localizing attributes on the desired items while minimizing unintended leakage. Furthermore, it performs the best in Precision, demonstrating its effectiveness in preventing attribute confusion.
In contrast, models such as SDXL~\cite{podell2023sdxl} and Multi-T2I-Adapter~\cite{mou2024t2i}, while strong in Recall, exhibit lower Precision scores, suggesting that attributes may inadvertently leak to unintended items, as shown in \cref{fig:qualitatives} (first row, the striped trousers). 
On the other hand, T2I-Adapter~\cite{mou2024t2i}, ControlNet~\cite{zhang2023adding}, and AnyControl~\citep{sun2024anycontrol} show relatively lower performance in both Precision and Recall. Models that underwent fine-tuning also improved performance, with Multi-T2I-Adapter~\cite{mou2024t2i} attaining the second-highest F1 score following \methodshort{}.
Finally, we measure the statistical relevance of these results and record a high Krippendorff's $\alpha = 0.81$, testifying to a high inter-annotator agreement.

\noindent\textbf{Qualitative results.}
We qualitatively analyze the attribute localization ability of different models, evaluating whether models accurately associate attributes with the intended garment, ensuring that details such as patterns, silhouettes, and structural elements are correctly rendered in their designated locations.
As shown in \cref{fig:qualitatives} (top), the input description specifies three main items associated with their attributes and garments. In particular, a plain single-breasted blazer jacket, a floral top, and striped pants. 
In this case, a proper generation should be able to faithfully follow the description, placing the floral pattern on the top while keeping the jacket's plain pattern.
As shown, \methodshort is the only approach that effectively generates a plain jacket with a floral top and striped pants. 
In contrast, multiple baselines and state-of-the-art models exhibit errors.
For instance, Multi-T2I-Adapter successfully follows the sketch, but fails to capture the floral pattern of the top. On the other hand, both IP-Adapter and T2I-Adapter wrongly localize the ``striped'' attribute to the jacket, failing to maintain semantic alignment with the prompt. These trends are not limited to isolated cases but are consistently observed across multiple examples, as denoted by the results in \cref{tab:vlm_stats} and \cref{tab:subjective}. More qualitatives are available in \suppmat.

\subsection{Ablation Analysis}\label{subsec:ablations}
In this section, we present an ablation analysis of the key components used in \methodshort, and evaluate the impact of different design choices, such as the image encoder, diffusion guidance strategy,
and the number of pooling tokens used in the Pair-Former module.

\noindent\textbf{Sketch Encoder.}
In the first section of \cref{tab:ablations}, we analyze different encoders for our sketch guidance. Training a dedicated sketch encoder end-to-end simultaneously to \methodshort (Trained) is ineffective, resulting in the lowest SSIM recorded ($.615$).
We hypothesize this is due to the pre-trained text features dominating the initial training phase, causing back-propagation to neglect the sketch encoder optimization.
Thus, to enhance sketch guidance, we test different frozen pre-trained image encoders, namely CLIP~\cite{radford2021learning}, BLIP2~\cite{li2023blip}, and DINOv2~\cite{oquab2023dinov2}.
DINOv2 (\methodshort) results in the highest SSIM ($.678$), with both CLIP and BLIP2 achieving subpar SSIM performance ($.623$ and $.630$ respectively).

\noindent\textbf{Diffusion Guidance.}
The second section of~\cref{tab:ablations} showcases different diffusion guidance strategies. In the first experiment (No Pooling), we compose our conditioning sequence $P$ by concatenation of all the $N$ paired features coming from the projectors without any pooling operation, \ie, $P = [h_1^S;h_1^T;\dots;h_N^S;h_N^T]$.
While this approach keeps all the encoder information available for guidance, the model ignores the sketch information, focusing only on textual conditioning, as testified by the high VQAScore ($.724$) and low SSIM ($.623$). We hypothesize this is due to the large number of uninformative tokens coming from the image encoder, \ie, the tokens where no sketch is present, pushing the model to ignore their guidance and focus only on the feature-rich textual features.
Averaging all the pairs into a single unified representation
$P = \frac{1}{N}\sum_{i=1}^{N}[h_i^S;h_i^T]$,
improves SSIM, but is unsatisfactory in textual adherence with a low $.676$ VQAScore.
By compressing the pairs' features and deferring the pair-merging operation to the diffusion process as presented in \cref{subsec:pairformer}, \methodshort achieves the best performance in both compositional semantic alignment ($.749$ VQAScore) and sketch-guidance ($.678$ SSIM).

\noindent\textbf{Pair-Former.}
We first evaluate the impact of our Pair-Former by replacing it with standard cross-attention layers (Cross-attn), resulting in lower performance across all metrics. We believe this is due to cross-attention layers limiting interactions to cross-modality only, while Pair-Former's self-attention enables both intra-/cross- modality interactions between text and sketch.
Additionally, to validate the feature compression performed by our Pair-Former, we augment the number of learning pooling tokens and find that,
interestingly enough, a higher number of pooling tokens (64 vs. 32) lowers the performance of \methodshort by $.08$ in VQAScore. We hypothesize this is due to both the increased number of trainable parameters and
the data redundancy introduced by the larger number of tokens, causing issues similar to the No Pooling ablation experiment.

\begin{table}
\centering
\resizebox{0.9\linewidth}{!}{%
    \begin{tabular}{c c ccc}
        \toprule
        \multirow{2}{*}{\textbf{Component}} & \multirow{2}{*}{\textbf{Choice}}& \multicolumn{3}{c}{\textbf{Compositional Alignment}}\\ 
        \cmidrule(lr){3-5}
         &  & LocalCLIP $\uparrow$ & VQAScore $\uparrow$ &  SSIM $\uparrow$\\
        \midrule
        \multirow{3}{*}{Sketch Encoder}
         & Trained & .762 & .689 & .615 \\
         & CLIP & .802 & .685 & .623\\
         & BLIP2 & .797 & .678 & .630\\
        \midrule
        \multirow{2}{*}{Diffusion Guidance}
        & No Pooling & .813 & .724 & .623 \\
        & Mean Pooling & \textbf{.819} & .676 & .659 \\
        \midrule
        \multirow{2}{*}{Pair-Former} & Cross-attn & .804 & .681 & .626 \\
        & 64 Tokens & .804 & .669 & .628 \\
        \midrule
        & \methodshort &  .813 & \textbf{.749} & \textbf{.678} \\
        \bottomrule
    \end{tabular}
}
    \caption{Ablation over different components of \methodshort}
  \label{tab:ablations}
\end{table}

\section{Conclusions}
\label{sec:conclusions}
In conclusion, we tackled the challenging problem of \textit{image generation with localized sketch-text} pairs, featuring a more realistic scenario in the fashion design process. We proposed a novel method \textit{\methodshort{}}, introducing a dynamic conditioning strategy that mitigates attribute confusion between input sketch-text pairs through modular processing of pairs and defers multi-condition fusion to the downstream denoising stage. We then integrate our approach with Stable Diffusion via an adapter-based method to effectively aggregate localized information. Moreover, we introduced \textit{\dataset{}}, a new dataset with high-quality fashion images associated with localized sketch-text pairs, to support model training and evaluation. 
\methodshort{} achieves state-of-the-art performance on \dataset{} via both quantifiable metrics and human evaluation, in terms of text-image alignment and visual grounding of local attributes.
Future works will explore more appropriate evaluation techniques for localized text-image alignment with fine details. Furthermore, \methodshort{} currently struggles with generating non-visually grounded attributes (\eg, ``summer-vibe''), likely due to CLIP’s limited sensitivity to abstract concepts~\citep{talon2025seeing}. As such, an interesting future direction would be to include more powerful textual encodings to include such abstract concepts.

\section{Acknowledgment}
This study was supported by LoCa AI, funded by Fondazione CariVerona (Bando Ricerca e Sviluppo 2022/23), PNRR FAIR - Future AI Research (PE00000013) and Italiadomani (PNRR, M4C2, Investimento 3.3), funded by NextGeneration EU.
This study was also carried out within the PNRR research activities of the consortium iNEST (Interconnected North-Est Innovation Ecosystem) funded by the European Union Next-GenerationEU (Piano Nazionale di Ripresa e Resilienza (PNRR) – Missione 4 Componente 2, Investimento 1.5 – D.D. 1058 23\/06\/2022, ECS\_00000043). This manuscript reflects only the Authors’ views and opinions. Neither the European Union nor the European Commission can be considered responsible for them.
We acknowledge ISCRA for awarding this project access to the LEONARDO supercomputer, owned by the EuroHPC Joint Undertaking, hosted by CINECA (Italy).
We acknowledge EuroHPC Joint Undertaking for awarding us access to MareNostrum5 as BSC, Spain.
Finally, we acknowledge HUMATICS, a SYS-DAT Group company, for their valuable contribution to this research.

{
    \small
    \bibliographystyle{ieeenat_fullname}
    \bibliography{main}
}

\clearpage
\maketitlesupplementary
\setcounter{page}{1}
\setcounter{figure}{4}
\setcounter{equation}{0}
\setcounter{section}{7}
\renewcommand{\thesection}{\AlphAlph{\value{section}-7}}

\section{Overview}
\label{sec:overview}
In this supplementary material, we supply more details regarding the implementation of our proposed method~\methodshort (\cref{supp:sec:method}). Then, we describe in more detail our subjective evaluation protocol with human evaluation (\cref{supp:sec:humaneval}).
In~\cref{supp:global} we explore the impact of different global descriptions during inference.
\cref{supp:data_recipe} provides additional details on the creation of the \dataset{} dataset.
Finally, we report more qualitative results to provide visual evidence of how our proposed method can improve image generation with a better visual grounding of localized textual attributes (\cref{supp:sec:qualitative}).

\section{More details on \methodshort{} implementation} \label{supp:sec:method}
\noindent\textbf{Implementation Details.} 
In \methodshort, the image and text projectors are linear layers followed by a layer normalization operation, while
the Pair-Former is implemented as a self-attention transformer with two self-attention blocks.
As diffusion backbone, we employ Stable Diffusion XL~\cite{podell2023sdxl}, using a generic global description as a prompt to contextualize the generation in the fashion domain while ensuring that all garment-level information comes from our adapter model. 
All the Stable Diffusion family models and weights, as well as cross-attention blocks and training utilities, were gathered from~\cite{von-platen-etal-2022-diffusers}.
During training, we freeze all model weights, except for the Image and Text projectors, our Pair-Former, and the additional cross-attention blocks.
\methodshort and all the fine-tuned approaches are trained on the train split of \dataset.

We train \methodshort following the standard Stable Diffusion procedure~\cite{rombach2022high}, while all other fine-tuned adapters are trained using their official implementations and default parameters. For \methodshort, we use Adam~\cite{kingma2014adam} optimizer, learning rate of $1e^{-5}$, and a total batch size of 32, while other approaches are trained using their default hyper-parameters.

\section{Details on human evaluation} \label{supp:sec:humaneval}
We design an attribute-focused questionnaire that focuses on the objective attribute existence, rather than subjective user preferences.
This is done to avoid introducing bias from human perception, which can be influenced by complex factors, \eg, image quality, that are irrelevant to localized attribute grounding. 
Specifically, we aim to evaluate whether an attribute is correctly localized in the generated image and whether it appears in unintended regions. 

In the study, images are generated using models from \cref{subsec:comparisons} with the same sketch and textual descriptions, 
and are then refined using the \texttt{Stable Diffusion XL Refiner}\footnote{https://huggingface.co/stabilityai/stable-diffusion-xl-refiner-1.0} model to avoid bias from the overall image quality.
Notably, we enrich each garment description with a pattern attribute during generation (\eg, striped, dotted) so that each garment is assigned a unique pattern, \ie the same pattern can't appear on more garments in the same picture. 
The rationale behind this decision is twofold: i) patterns are one of the few attributes that can be applied to any kind of garment without limitations, as opposed, for instance, to sleeve length or neckline shape; ii) by being uniquely assigned and easily identifiable, user can detect attribute confusion with relative ease. 

In practice, we instruct the user to answer a pair of templated questions, \texttt{``Consider the garment <class1>: is it <attribute>?''}, and \texttt{``Consider the garment <class2>: is it <attribute>?''}, where \texttt{<class>} is the garment class, and the \texttt{<attribute>} is an attribute that is visually noticeable, \eg, ``check" or ``striped".
Since only one item per image can contain the sampled attribute, 
if it fails to represent an item or attribute or binds it to an unrelated garment, the responses will reflect this misalignment. 

To measure this quantitatively, we use \textit{Recall}, \textit{Precision}, and \textit{F1 scores} as metrics.
Recall is defined as the fraction of times that a specified attribute is correctly applied to the intended clothing garment. In other words, Recall measures how often the model successfully localizes the desired attribute on the correct garment. 
Formally, Recall is computed as:
\begin{equation}
\text{Recall} = \frac{\text{TP}}{\text{TP} + \text{FN}}
\end{equation}
where:
\begin{itemize}
    \item TP (True Positives): The attribute is reflected correctly at the intended garment in the generated image.
    \item FN (False Negatives): The attribute should have been reflected, but is missing.
\end{itemize}

Similarly, Precision is defined as the extent to which the generated attribute appears exclusively on the intended item without being mistakenly applied to other objects:
\begin{equation}
\text{Precision} = \frac{\text{TP}}{\text{TP} + \text{FP}}
\end{equation}
where a False positive (FP) is defined as when an attribute appears on unintended garments indicating attribute confusion.

A high recall value indicates that attributes are generated where they are supposed to, whereas a high precision score indicates that attributes are not leaking to irrelevant objects.
F1 Score offers a balanced view of both precision and recall, taking into account both false positives and false negatives:
\begin{equation}
\text{F1 Score} = 2 \times \frac{\text{Precision} \times \text{Recall}}{\text{Precision} + \text{Recall}}
\end{equation}

\section{Impact of global description on generation}\label{supp:global}
\begin{figure}
    \centering
    \includegraphics[width=0.83\linewidth]{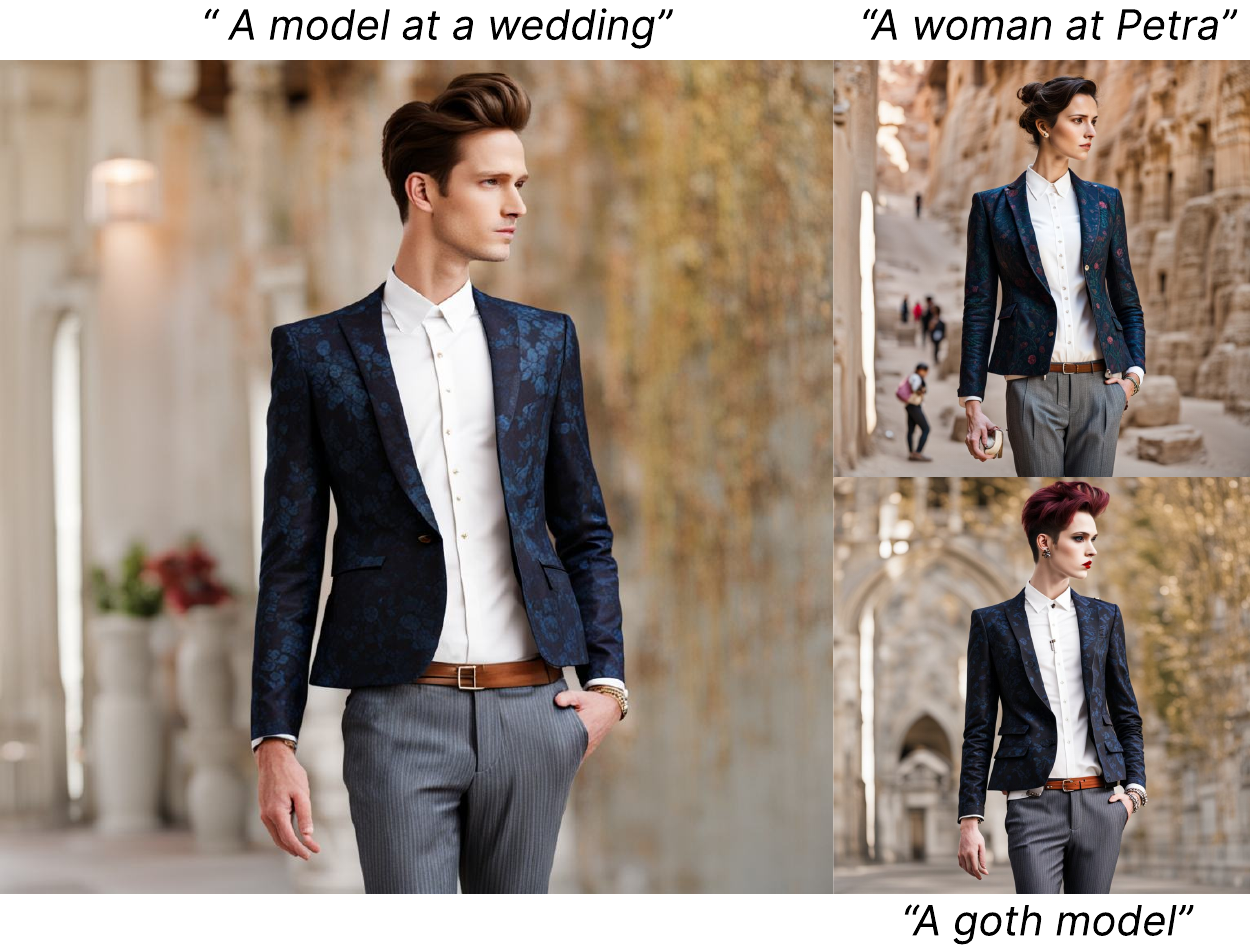}
    \caption{Effects of different global descriptions on the generation of \methodshort. By changing the global description, we are able to customize general aspects such as the background and style of the model and the outfit (see text).}
    \label{fig:qualitatives_for_qual}
\end{figure}

To illustrate the impact of our global conditioning 
$T_g$, we present in \cref{fig:qualitatives_for_qual} three examples where the local sketch and text remain fixed, while the global text varies across three different descriptions.
Notably, the description ``A goth model'' results in an image with extra stylistic details, such as additional bracelets, earrings, pale skin, and red-tinted hair, while maintaining the overall properties of the garments.

\section{Data Recipe for Sketchy}\label{supp:data_recipe}
In this section, we add specific details regarding the construction of \dataset. Please note that our pre-computed data and data curation scripts are available on our project website.
First, we organize the garments in Fashionpedia's annotations~\cite{jia2020fashionpedia} to create our localized hierarchical structure (\cref{sec:app-hierarchical}). Then, we use an LLM to automatically generate a natural language description of the whole-body items starting from their attributes (\cref{sec:app-partial-desc}). Finally, we generate a sketch for every item that appears in each image (\cref{sec:app-sketch}).
\subsection{Local garments organization}
\label{sec:app-hierarchical}
As stated in the main document, Fashionpedia contains annotations for both whole-body items, \eg, shirts, and garment-parts, \eg, sleeves. Despite this, Fashionpedia does not explicitly link garment-parts with their respective whole-body item, \eg, the shirt and sleeve annotations are not linked to each other. This can become problematic when multiple suitable whole-body items appear in a single image for a given garment-part: if we have a ``sleeve'' annotation, and both a ``shirt'' and ``jacket'' in the same image, it is not clear which item the sleeve belongs to.
At the same time, garment-part annotations contain fine-grained attributes that we would like to associate with the whole-body item, \eg, the ``long'' attribute for the sleeves does not appear in the whole-body shirt annotation.

To amend this, we use the segmentation masks provided by Fashionpedia to find associations between garment-parts and whole-body items. 
Specifically, 
for images containing garment parts, we calculate the overlapping area between each garment-part mask and the masks of whole-body items. The garment-part’s attributes are then assigned to the whole-body category with the largest overlapping area. In extreme cases where no overlap exists, we assign the garment attribute to the whole-body category that is most frequently associated with this specific garment part across the dataset, \ie, based on co-occurrence statistics. 

\subsection{Partial Descriptions}
\label{sec:app-partial-desc}
Fashionpedia annotations contain a list of attributes describing the properties of the item depicted (for both whole-body and garment parts). While these lists of attributes could be used directly to condition a text-to-image model, we found the interaction to be unrealistic: in a real-world case, we believe the user would rather use natural language descriptions to build a coherent sentence describing the contents of the image.
For this reason, for each whole-body annotation we use an off-the-shelf \texttt{Llama 3.1 8B-Instruct}\footnote{https://huggingface.co/meta-llama/Llama-3.1-8B-Instruct} model to generate a natural language description starting from the attribute list.
The used prompt is reported in~\cref{sys-p}.
To further guide the model in the generation, we provide some in-context-learning examples, describing both the input structure and corresponding expected output, as depicted in~\cref{icl-p}.

\begin{figure*}
\begin{codeblock}
    \begin{verbatim}
# SYSTEM PROMPT
"You are a fashion expert.  Describe the clothing item concisely based on 
the information provided, strictly within 70 tokens.
You need to prioritize keeping the information that may influence 
the appearance the most, while trying to describe the image 
as informatively as possible.
Within each whole-body item, there could be sub-items, each may have 
its own descriptive attributes.

Here is the structure of the clothing item information provided:
The item information is a structured dictionary including the following keys:
(1) "category": A string indicating the main item category 
(e.g., "Coat", "Trousers").
(2) "top_level": A list [] of attributes that provide a general description 
of the main item (e.g., ["long", "wool"]).
(3) "sub_level": A list [] of dictionaries {} where each dictionary describes 
a specific part of the main item. Each dictionary contains:
- The part's name as the key (e.g., "Collar", "Pockets"). 
- A list [] of attributes as the value that 
describes the details of that part (e.g., ["wide", "deep", "large"]).

Please provide a cohesive description of the item, incorporating all 
the details provided for both the main item and its sub-items. 
Ensure the description maintains clarity and preserves the hierarchical 
relationship between main items and sub-items.
Refrain from giving any personal opinion. 
You must reply in the format of a Python dictionary {desc: description}."

\end{verbatim}
\end{codeblock}
\caption{System prompt for generating whole-body garment descriptions from attributes.}\label{sys-p}
\end{figure*}

\begin{figure*}
\begin{codeblock}
    \begin{verbatim}
# IN-CONTEXT SAMPLE 1
# input structure
"{
    "category": "coat",
    "top_level": ["long", "wool"],
    "sub_level": [
        {"Collar": ["wide"]},
        {"Pockets": ["deep"]},
        {"Buttons": ["large"]}
    ]
}"
# output
"{desc: A long wool coat with a wide collar, deep pockets and large buttons}"

# IN-CONTEXT SAMPLE 2
# input structure
"{
    "category": "trousers",
    "top_level": ["slim-fit"],
    "sub_level": [
        {"Stitching": ["subtle"]},
        {"Leg": ["tapered"]}
    ]
}"
# output
"{desc: Slim-fit trousers with subtle stitching and a tapered leg}"

# IN-CONTEXT SAMPLE 3
# input structure
"{
    "category": "shirt",
    "top_level": ["cotton"],
    "sub_level": []
}"
# output
"{desc: A cotton shirt}"

# IN-CONTEXT SAMPLE 4
# input structure
"{
    "category": "shoe",
    "top_level": [],
    "sub_level": []
}"
# output
"{desc: A pair of shoes}"
    \end{verbatim}
\end{codeblock}
\caption{In-context-learning samples appended to the system prompt for every input annotation for generating whole-body garment descriptions from attributes.}\label{icl-p}
\end{figure*}

\subsection{Partial Sketches}
\label{sec:app-sketch}
To generate the localized sketches we rely on our hierarchical annotation structure and an off-the-shelf Image-to-Sketch model, \texttt{Photo-sketching}~\cite{li2019photo}.
Photo-sketching takes an image as input and generates the corresponding human-like sketch. 
However, our goal is to generate localized sketches, \ie, sketches of single items inside the image, and not the entire scene.
To do so, we first obtain the segmentation mask (from the Fashionpedia annotations) of every whole-body item inside the input image. 
Then, for every item, we crop the image around the entire mask and resize the image, effectively ``zooming in'' on the item of interest. The Photo-sketching model is then used to generate the sketch of the cropped image. Finally, we remove unwanted background information by masking the generated sketch with the item segmentation mask.

Thanks to our zooming-in operation, we found that Photo-sketching was able to focus on more fine-grained details, such as pockets and patterns, that would have otherwise been lost by generating a sketch of the entire image.
At the same time, our final masking operation removes unwanted information coming from neighboring items and background objects. We found that this allows the generative model downstream (\methodshort) to focus on the item shape during training, avoiding information corruption coming from unrelated items being present in the sketch.

\section{More qualitative results} \label{supp:sec:qualitative}
\begin{figure*}[t!]
    \centering
    \includegraphics[width=0.95\textwidth]{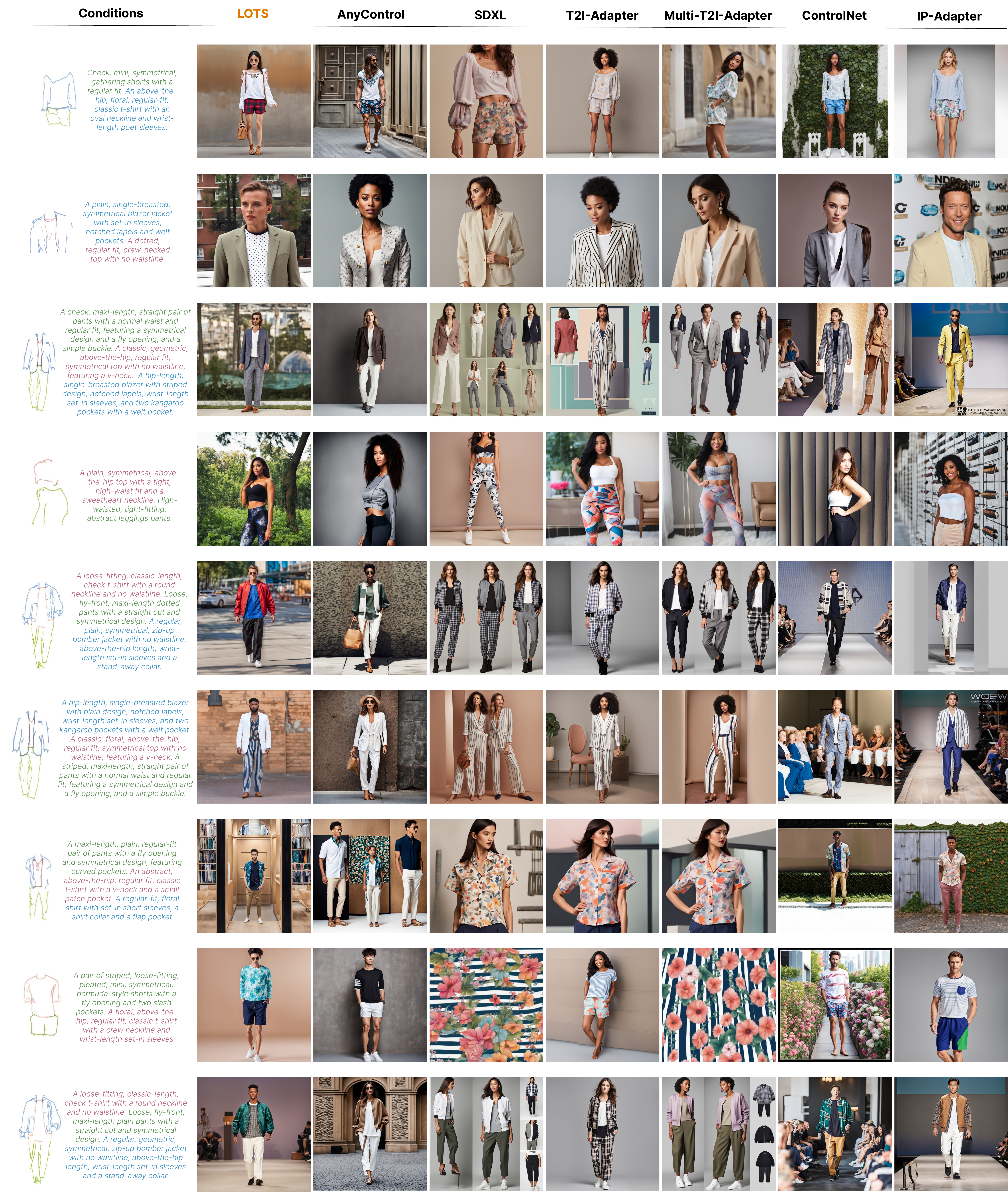}
    \caption{More qualitative results of \methodshort{} in comparison with baselines and competitors as described in \cref{sec:experiments}. In this table, SDXL~\cite{podell2023sdxl}, AnyControl~\citep{sun2024anycontrol} are zero-shot approaches, T2I-Adapter~\cite{mou2024t2i}, Multi-T2I-Adapter~\cite{mou2024t2i}, ControlNet~\cite{zhang2023adding}, IP-Adapter~\cite{ye2023ip} are fine-tuned versions. Given paired localized text-sketch are conditioning inputs, \methodshort{} can more correctly reflect fine-detailed attributes in the intended local region in the generated images, effectively mitigating attribute confusion, a common problem when only a global description is provided as the textual condition.}
    \label{fig:qualitatives_supp}
\end{figure*}

\begin{figure*}[t!]
    \centering
    \includegraphics[width=0.95\textwidth]{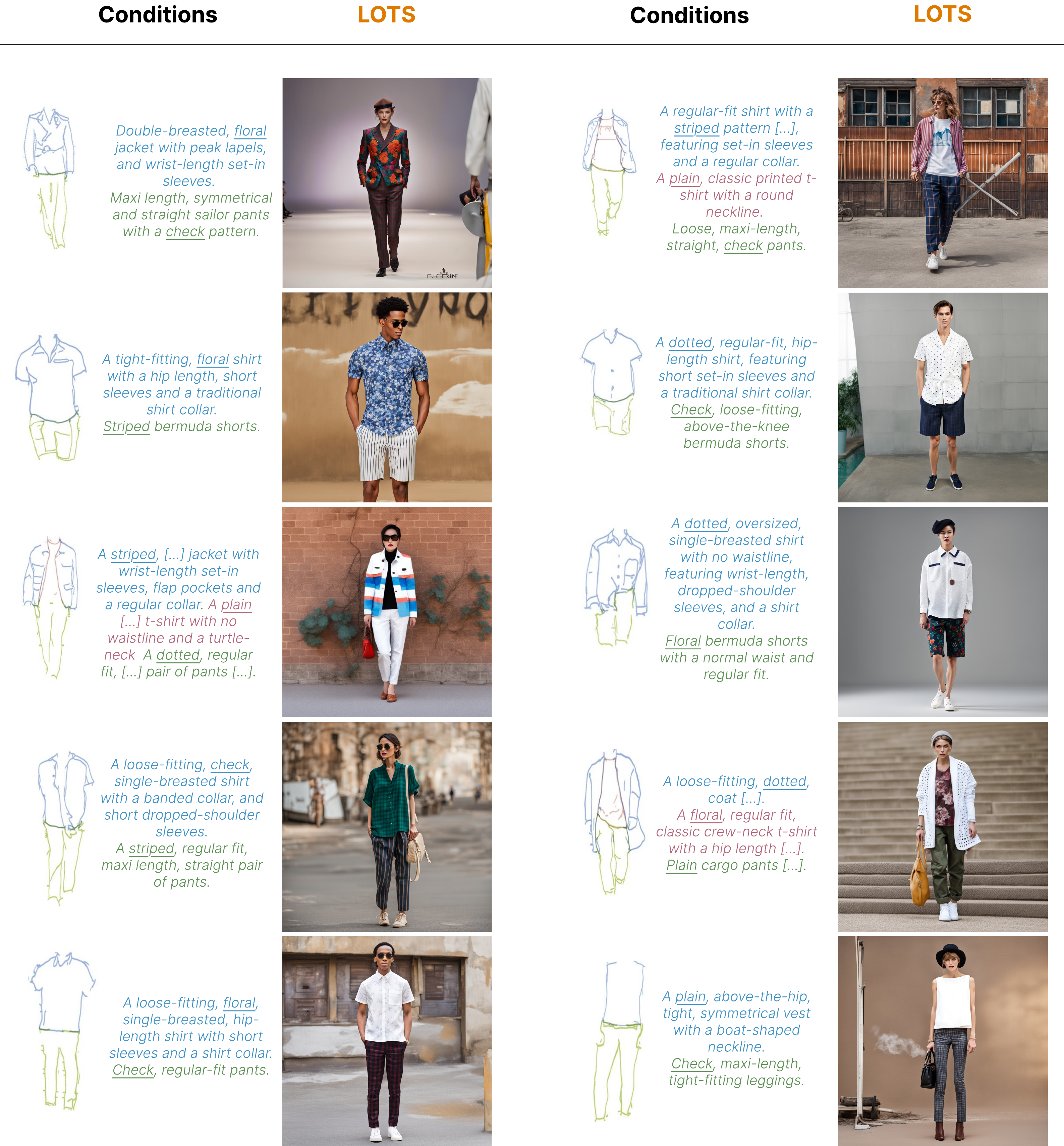}
    \caption{More qualitative results of \methodshort{}. We underline the textual pattern attributes to facilitate visual inspection.}
    \label{fig:more_qual}
\end{figure*}

\Cref{fig:qualitatives_supp} shows more qualitative results of \methodshort{} in comparison with baselines and competitors as described in \cref{sec:experiments}. 
Given paired localized text-sketch as conditioning inputs, \methodshort{} can more correctly reflect fine-grained attributes in the intended local region in the generated images, effectively mitigating attribute confusion, a common problem when only a global description is provided as the textual condition. 
We observe two main failure cases of state-of-the-art approaches. On the one hand, models tend to confuse attributes, conveying them on main whole-body garments (see striped and checked outfits in SDXL, T2I-Adapter, and Multi-T2I-Adapter). This is in line with the intuition that the model focuses on the most peculiar attribute in the single global description. On the other hand, generation can completely fail to follow the conditioning sketch, \eg, three out of seven generations for Multi-T2I-Adapter, indicating that explicit merge of the spatial conditioning, before generation, could possibly corrupt the guiding information. In contrast, with the modularized processing of pairs and the diffusion pair guidance, which overcomes the pooling of conditioning localized information, \methodshort{} allows for effective and fine-grained conditioning at both semantic (textual descriptions) and spatial level (sketches).
More results in \cref{fig:more_qual}.

\end{document}